# FPMT: Enhanced Semi-Supervised Model for Traffic Incident Detection

Xinying Lu and Jianli Xiao*

School of Optical-Electrical and Computer Engineering
University of Shanghai for Science and Technology, Shanghai, China, 200093
`222240474@`st.usst.edu.cn(Xinying Lu), audyxiao@sjtu.edu.cn(Jianli Xiao)

**Abstract.** For traffic incident detection, the acquisition of data and labels is notably resource-intensive, rendering semi-supervised traffic incident detection both a formidable and consequential challenge. Thus, this paper focuses on traffic incident detection with a semi-supervised learning way. It proposes a semi-supervised learning model named FPMT within the framework of MixText. The data augmentation module introduces Generative Adversarial Networks to balance and expand the dataset. During the mix-up process in the hidden space, it employs a probabilistic pseudo-mixing mechanism to enhance regularization and elevate model precision. In terms of training strategy, it initiates with unsupervised training on all data, followed by supervised fine-tuning on a subset of labeled data, and ultimately completing the goal of semi-supervised training. Through empirical validation on four authentic datasets, our FPMT model exhibits outstanding performance across various metrics. Particularly noteworthy is its robust performance even in scenarios with low label rates.

**Keywords:** Traffic Incident Detection, Semi-supervised Learning, Generative Adversarial Networks.

## 1 INTRODUCTION

In the realm of intelligent traffic systems, traffic incident detection refers to accurately identifying unpredictable incidents such as traffic accidents, road maintenance, and severe congestion in terms of both location and time[1]. Ensuring the efficient operation of urban traffic and enhancing the safety of people's travel constitute one of the core functionalities of intelligent traffic systems, involving the automatic detection of traffic incidents. This involves promptly identifying and addressing these incidents to improve overall traffic flow. However, relying on traffic flow data for incident detection necessitates continuous data collection and labeling by professionals, which is resource-intensive. Current research has predominantly focused on deep learning methods, which typically require substantial labeled data for training, presenting a significant challenge in applying deep learning to traffic incident detection with limited labeled data [2].

To address the scarcity of label information in real-world scenarios due to difficulty in acquisition, semi-supervised learning has garnered attention. Semi-supervised learning successfully alleviates the over-reliance on supervised information in machine learning, leveraging unlabeled samples to train a reliable classifier for effective predictions in target categories based on the popular assumption and clustering hypothesis. Various semi-supervised learning models employ different strategies for handling unlabeled samples, such as entropy minimization, consistency regularization, and data augmentation. Examples include the Mean Teacher model [3] and Virtual Adversarial Training (VAT) [4]. However, these semi-supervised learning methods are based on the assumption that the distribution of labeled and unlabeled data pairs is entirely identical, treating labeled or unlabeled data separately [5].

Due to the limited availability of labeled data during the training process of semi-supervised learning, overfitting is prone to occur. To better utilize unlabeled data, Bertholot et al. proposed the MixMatch method [6]. This method generates mixed samples by interpolating different samples through MixUp and interpolates mixed pseudo-labels for different samples. The authors also introduced the FixMatch model [7], which achieves state-of-the-art performance in semi-supervised learning benchmarks by using weakly augmented unlabeled images to generate high-confidence pseudo-labels and training the model with strongly augmented image versions. However, these interpolation methods are designed for image data, and discrete text data, interpolation needs to be performed in the corresponding hidden space, leading to the development of a new semi-supervised learning method for text data called MixText [8].

To further explore the latent information in unlabeled data by utilizing more rational training, data augmentation, and loss calculation strategies, integrating their complementary strengths, a semi-supervised traffic incident detection model is proposed based on the MixText framework. In the data augmentation module, as traffic incident datasets often exhibit significant imbalances and small scales, the application of GANs is proposed to balance and expand the dataset. A probability pseudo-mixing strategy is employed in the hidden space when performing Mix-up, assigning confidence to samples entering the mixture, thereby giving more weight to samples with higher confidence to enhance regularization. In terms of training strategy, unsupervised training is initially performed on all data, followed by supervised training on a subset of labeled data, and ultimately, semi-supervised fine-tuning is conducted to improve detection rates. In the semi-supervised fine-tuning phase, pseudo-labels are first predicted for unlabeled data, and confidence is assigned to these pseudo-labels. When labeled and unlabeled data enter the model's hidden layer, interpolation is performed based on the confidence ratio, iterating through training to extract latent information and obtain classification results. The model is experimentally validated on four real datasets and compared with baseline models. Through ablation studies, the effectiveness of each module is demonstrated. The results show that the proposed semi-supervised traffic incident detection model FPMT performs exceptionally well with very limited labeled data.

In summary, the main contributions of this paper are as follows:
- Proposing a novel semi-supervised traffic incident detection model that exhibits outstanding performance in scenarios with extremely low label rates.

- Introducing GANs in the data augmentation module to balance and expand the dataset.
- Optimizing the interpolation strategy in the hidden layer under the MixText framework to enhance regularization.
- Applying a training strategy that involves supervised training initially, followed by semi-supervised fine-tuning to improve detection rates.
- Conducting extensive experiments on four real datasets, demonstrating the effectiveness of the proposed semi-supervised traffic incident detection model and validating the effectiveness of each module.

## 2   Related Work

The development of Traffic Incident Detection traces back to 1965 when the California algorithm [9] utilized fluctuations in upstream and downstream traffic flow data to identify the occurrence of incidents. Subsequently, the standard deviation algorithm [10] employed standard deviation values to observe the average trends of preceding time intervals and the current transformation trends, thereby discerning whether a traffic incident has occurred. Following this, Bayesian algorithms [11], rooted in statistical theory, sequentially emerged. However, they exhibited an overreliance on past experiences, posing flexibility challenges. With the robust growth of machine learning post-1990, classical models such as Support Vector Machines (SVM) [12] and Random Forests [13] were applied to this task. Various artificial intelligence algorithms found applications in the field of Traffic Incident Detection, including Convolutional Neural Networks (CNN) [14] and Long Short-Term Memory Neural Networks [15] as part of deep learning methods.

The development of semi-supervised learning commenced in 2005 when Grandvalet and others proposed the entropy minimization method [16], becoming the most classic and commonly used deep semi-supervised learning algorithm and strategy. This approach effectively integrates unlabeled data in semi-supervised learning, demonstrating robust performance, especially when addressing violations of generated model error specifications or "cluster assumptions." Subsequent developments in semi-supervised learning models are intricately tied to four aspects: entropy minimization, consistency regularization, data augmentation, and pre-training fine-tuning.

### 2.1   Consistency Regularization

In 2018, Tarvainen et al. introduced the Average Teacher Model based on Consistency Regularization [3], significantly improving performance compared to previous methods. In the same year, Miyato et al. proposed the Virtual Adversarial Training method [4], which involves computing the gradient of the network to generate adversarial samples. These adversarial samples are designed to maximize the network's vulnerability, and by combining them with pseudo-labels derived from the original samples, the network can be correctly trained, maximizing its robustness against interference. In 2020, Sohn et al. presented the FixMatch method [7], which involves applying slight transformations to unlabeled samples for initial predictions and selecting samples with high confidence to assign pseudo-labels. Subsequently, these samples undergo more substantial transformations, and the consistency loss is computed between the pseudo-labels and the blurred predictions after strong transformations, thereby enhancing the learning effectiveness of the network.

### 2.2    Entropy Minimization

In 2013, Lee proposed the Pseudo-Label method [17], which has become the most widely used semi-supervised learning approach. Essentially, it also leverages the strategy of entropy minimization employed by the network predictions. The Pseudo-Label method primarily involves selecting samples with high confidence during the learning process, transforming their network predictions into pseudo-labels corresponding to the class with the highest predicted probability. These pseudo-labels are then utilized to assist in the network training process. In 2016, Laine and Aila introduced the Π−model [5]. This method utilizes two structurally identical but parametrically distinct network models, aiming for both networks to produce the same predictions for identical samples. Consequently, when one network generates incorrect labels for unlabeled samples during training, the other network can correct them. This consistency training strategy avoids the robustness issues associated with the aforementioned entropy minimization strategy.

### 2.3    Data Augmentation

In 2019, Wang et al. proposed a straightforward yet effective semi-supervised learning method called Augmented Distribution Alignment [18]. This method employs adversarial training and interpolation strategies to alleviate sampling biases arising from limited labeled samples in semi-supervised learning. It aligns the empirical distributions of labeled and unlabeled data. In the same year, Bertholot et al. introduced a novel semi-supervised learning model named MixMatch [6]. By unifying current mainstream semi-supervised learning methods, this model infers low-entropy labels on augmented, unlabeled examples and utilizes MixUp technology to blend labeled and unlabeled data. Cai et al. presented Semi-ViT [19], another semi-supervised learning model, introducing a probability pseudo-mixing mechanism for interpolating unlabeled samples and their pseudo-labels, enhancing the regularization effect.

### 2.4    Pretraining and Fine-tuning

In 2018, Howard et al. introduced Universal Language Model Fine-tuning (ULMFiT) [20], incorporating key techniques for fine-tuning language models. In 2020, Ting et al.'s SimCLR[21] model demonstrated a significant improvement in accuracy when fine-tuning on only 1% of labels. Subsequent research utilized SimCLRv2 [22] for unsupervised pre-training of a large ResNet model, followed by supervised fine-tuning on a small set of labeled examples. Knowledge from unlabeled examples was distinguished to enhance and transfer task-specific knowledge.

## 3  Method

### 3.1  Fusion of Training Pipeline

The paradigm shift in the training pipeline has made significant strides in improving model performance in recent years. For instance, in the FixMatch framework, the pipeline has been altered to first undergo unsupervised pre-training followed by self-supervised training fine-tuning. Similarly, in the SimCLRv2 framework, the approach involves initial unsupervised pre-training followed by supervised fine-tuning, ultimately employing knowledge distillation and transfer from unlabeled samples. In the training process of this study, following experimentation and exploration, a methodology akin to SimCLRv2 was adopted. Specifically, the decision was made to first conduct unsupervised pre-training on the entire dataset, then perform supervised fine-tuning on a subset of labeled data, and finally engage in semi-supervised training on both labeled and unlabeled data. Within the semi-supervised training framework, Probability MixText (PMT) was employed, incorporating probabilistic pseudo-mixing and GANs-based data augmentation techniques into the foundational MixText framework.

### 3.2  Probabilistic Pseudo Mixup

For the proposed data augmentation technique Mixup applied to image data, linear interpolation is performed at the pixel level of the input images. Specifically, it involves blending the pixel values of the original images in a certain proportion. Correspondingly, the labels are mixed in the same ratio, resulting in new samples and labels with blended features. The mixing ratio $\lambda$ is derived from a Beta distribution, typically involving the random selection of two different samples, $x_q$ and $x_p$, along with their corresponding labels from the dataset $X = \{x_1, ..., x_m\}$ and labels $Y = \{y_1, ..., y_m\}$. Formally, for $p, q \in [0, m]$, the mixing process is defined as follows:

$$\tilde{x} = \lambda x_q + (1-\lambda) x_p, \tag{1}$$

$$\tilde{y} = \lambda y_q + (1-\lambda) y_p. \tag{2}$$

However, due to the varying qualities of data and pseudo-labels generated, the simple random selection of the mixing ratio $\lambda$ from the Beta distribution for poorly performing samples might lead to an undesired impact. This randomness could potentially allow low-quality data to influence high-quality data and affect loss calculations. To address this issue, the concept of probabilistic pseudo-mixing [19] is introduced.

Despite the lower quality of the data, it still holds valuable information. Probabilistic pseudo-mixing continues to involve random mixing of unlabeled data, but the mixing ratio $\lambda$ is no longer randomly generated from a Beta distribution. During the semi-supervised training phase, pseudo-labels and corresponding losses are generated. The confidence is determined based on the loss information from the two samples involved in the mixing. Through this mechanism, samples with higher confidence have a higher proportion in the mixed samples, and consequently, the pseudo-labels have a higher proportion. This weighting allows higher confidence samples to contribute more significantly to the semi-supervised loss calculation. This mixing strategy enhances regularization and provides greater flexibility.

### 3.3 Data Augmentation

To address the high imbalance and insufficient scale of traffic incident data, in the data augmentation module, the decision was made to employ Generative Adversarial Networks (GANs) to tackle these challenges. GANs are a common data augmentation technique capable of simulating the distribution of input data and capturing latent information to generate highly similar new data. Typically, GAN models consist of a generator and a discriminator, aiming to train the generator in such a way that the discriminator maximizes the probability of erroneously classifying generated samples as real samples. Previous research [23] has demonstrated that GANs produce high-quality data and have significantly contributed to advancements in various research domains.

### 3.4 MixText

The Probability Pseudo-Mixing (Tmix) technique was originally designed for image data and may not be directly applicable to text data due to its discrete nature. Therefore, interpolation is performed in the hidden space. In a BERT model with $H$ layers, the process involves selecting $x$ and $x'$ from the dataset and inputting them into the first layer to obtain their hidden representations, denoted as $h$ and $h'$, respectively. Then, at an intermediate layer, denoted as layer $E$, the hidden representations $h_E$ and $h'_E$ of these two samples are mixed using the Mixup operation, generating a new sample $h_m$ based on a random number $\lambda$ drawn from a Beta distribution for each batch. The formula for obtaining the new sample $h_m$ is as follows:

$$h_m = \lambda h_E + (1-\lambda) h'_E, \qquad (3)$$

simultaneously, the samples $x$ and $x'$ are input into the BERT model to obtain their corresponding pseudo-labels $y$ and $y'$, respectively. With the previously generated mixing ratio $\lambda$, the pseudo-label for the new sample is calculated as follows:

$$y_m = \lambda y + (1-\lambda) y'. \qquad (4)$$

This constitutes the framework of Tmix. Extending from this, MixText incorporates both labeled dataset $X_L$ and unlabeled dataset $X_U$ into the model training pro-

FPMT: Enhanced Semi-Supervised Model for Traffic Incident Detection    7cess. For labeled data, the model is trained using the supervised loss function, which is the cross-entropy loss:

$$L_x = -\frac{1}{N}\sum_{i=1}^{N}\sum_{j=1}^{C} y_{i,j} log(p_{i,j}), \quad (5)$$

where, $N$ denotes the batch size, $C$ represents the number of classes, $y_{i,j}$ signifies the $j$-th element in the true label of sample $i$, and $p_{i,j}$ is indicative of the predicted probability by the model for the $j$-th class of sample $i$.

For unlabeled samples, the Kullback-Leibler Divergence Loss (KL Divergence Loss) is employed as the consistency loss. This helps ensure that the model produces similar outputs for similar inputs, enhancing the model's consistency $L_u$. The loss calculation formula is as follows:

$$L_u = D_{KL}(softmax(outputs_u) \| targets_u), \quad (6)$$

where $outputs_u$ is the model's output for unlabeled data, and $targets_u$ is the pseudo-label calculated based on the predicted probabilities $p_{i,j}$.

For mixed data, when both mixed samples come from the labeled dataset, the model is trained using the supervised loss. When both mixed samples come from the unlabeled dataset, the model is trained using KL Divergence Loss. When one mixed sample comes from the labeled dataset and the other from the unlabeled dataset, the model is trained using both the supervised loss and KL Divergence Loss. The formula is as follows:

$$L = L_x + w \cdot L_u, \quad (7)$$

where $w$ is a weight used to balance the contributions of labeled and unlabeled samples.

The training process involves iterative mixing of labeled and unlabeled data with a certain probability, calculating the corresponding losses in each iteration.

### 3.5   FPMT

Building upon the MixText framework, the new semi-supervised traffic incident detection model, named FPMT, integrates the training strategy as a Fusion of Training Pipeline, the mixing strategy as a Probabilistic Pseudo Mixup, and the data balancing and augmentation strategy as GANs.

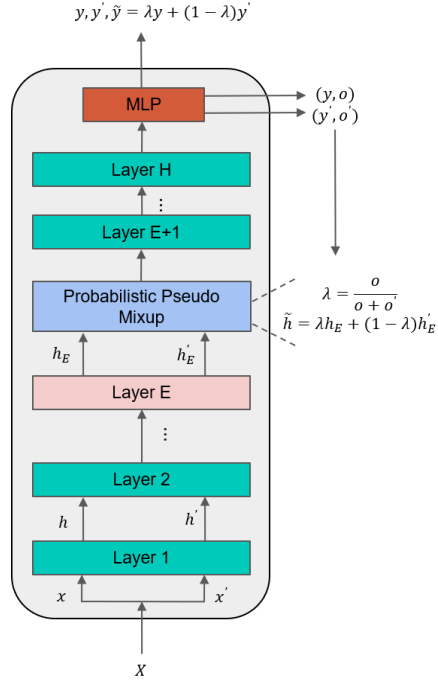

**Fig. 1.** Architecture of PTmix.

The framework of the Probability Pseudo Mixup model (PTmix) within FPMT is illustrated in Figure 1. Samples $x$ and $x'$ are input into PTmix, and the output layer provides their pseudo-labels and confidence scores $(y, o)$ and $(y', o')$. The mixing ratio $\lambda$ is determined based on the confidence proportions using the formula 8. Subsequently, based on the values of $\lambda$ and $1-\lambda$, samples $x$ and $x'$ are mixed at the layer $E$ of the model, producing hidden representations $h_E$ and $h'_E$ for the new sample, as well as mixing the pseudo-labels $y$ and $y'$ for the unlabeled data, resulting in the new sample $\tilde{h}$ and $\tilde{y}$.

$$\lambda = \frac{o}{o+o'}, \tag{8}$$

$$\tilde{h} = \lambda h_E + (1-\lambda) h'_E, \tag{9}$$

$$\tilde{y} = \lambda y + (1-\lambda) y'. \tag{10}$$

The semi-supervised fine-tuning stage of the FPMT model follows the framework of the PTmix model, as depicted in Figure 2. Initially, the dataset $X_O$ undergoes data augmentation using GANs to balance and expand the dataset. The augmented dataset is then partitioned into the labeled dataset $X_L$ and the unlabeled dataset $X_U$. Both $X_L$ and $X_U$ are fed into PTmix, generating predicted labels $Y_L$ for $X_L$, predicted labels $Y_U$ for $X_U$, and predicted labels $Y_M$ for mixed data $X_M$. The mixing strategy employed is probabilistic pseudo-mixing. Different loss functions are applied for calculating losses on different types of data, following the strategy outlined in Mix-Text, as described in Section 3.4.

Additionally, the FPMT model adopts a training strategy involving initial unsupervised training on all data, followed by supervised fine-tuning and ultimately semi-supervised fine-tuning.

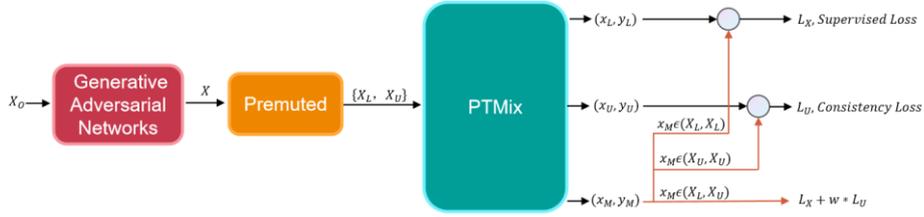

**Fig. 2.** Overall architecture of the proposed FPMT model.

## 4    Experiment

### 4.1    Datasets

To evaluate the proposed semi-supervised traffic incident detection model, four real-world datasets were utilized. These datasets include PeMS [24], I-880[25], Whitemud Drive [26], and NGSIM [27]. PeMS is a California-specific traffic flow database, that collects real-time data from over 39,000 independent detectors. It includes parameters of traffic flow, incident data, and weather information. I-880 originates from the renowned I880 highway traffic incident database in the United States, documenting traffic flow speed, occupancy data, and incident information on a 9.2-mile stretch of the highway. Whitemud Drive is a 28-kilometer-long highway in Edmonton, Alberta, Canada, equipped with circular detectors on main lanes and ramps to gather traffic parameters. NGSIM, initiated by the United States Federal Highway Administration, gathers real-time vehicle trajectory data for driving behavior analysis, traffic flow analysis, microsimulation modeling, and vehicle trajectory prediction. These datasets provide valuable information for traffic flow prediction, model analysis, and urban traffic planning and management.

### 4.2    Comparing Method

To validate the effectiveness of FPMT, it was compared with several recent models during the experimental phase. Among these, BERT [28] is a bidirectional encoder representation model that achieves significant performance improvements across various natural language processing tasks by jointly pretraining on the left and right context of the text, without requiring extensive task-specific architecture modifications during fine-tuning. VAT [4] is a regularization method based on virtual adversarial loss, achieving high performance in semi-supervised learning tasks by measuring the local smoothness of the input conditional label distribution. UDA [29] is a new approach in semi-supervised learning that employs advanced data augmentation methods such as RandAugment and back-translation, replacing simple noise operations and significantly improving performance across six languages and three visual tasks. DSP [30], by guiding the teacher to generate more accurate pseudo-labels through student feedback and combining consistency regularization, significantly improves text classification performance.

### 4.3    Experimental Setting

As traffic incident detection is a binary classification task, the class parameter is fixed at 2. For the layer selection of probabilistic pseudo-mix-up in PMT, it was observed that the mixing performed better at the 9th layer after training PMT separately. The model's decoder is based on Bert-base-uncased, and the output is classified through an additional linear layer. The learning rate for the BERT model's encoder is set to 0.00001, and the learning rate for the additional linear layer is set to 0.001. During the semi-supervised fine-tuning phase, for each dataset, GANs are utilized to balance and augment the dataset. For the augmented dataset, in each category, the number of unlabeled samples is set to 5000, while the number of labeled samples is set to 50, 100, and 1500, achieving label rates of 1%, 2%, and 30%, respectively.

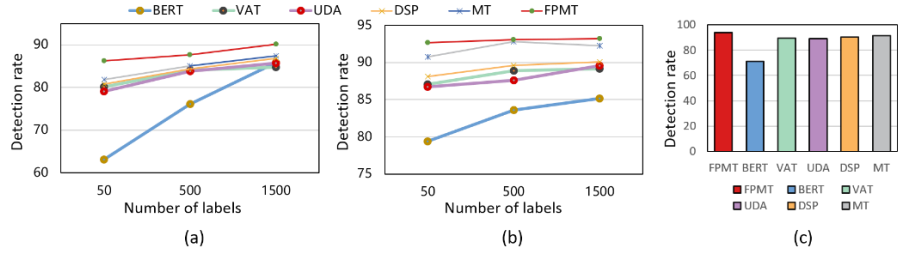

**Fig. 3.** (a) and (b) compare our FPMT model with baselines using different numbers of labeled samples (50, 100, 1500). Meanwhile, (c) represents the comparison with a fixed number of labeled samples set at 50.

### 4.4    Result

In the experiments, the selected evaluation metrics include Classification Rate (CR), Detection Rate (DR), and F1-score. After fixing the number of unlabeled samples for each category at 5000, experiments were conducted on four real datasets with varying numbers of labeled samples (50, 100, and 1500 for each category), and the results are presented in Table 1 and Figure 3. Our FPMT model achieved superior performance compared to the contrasted models, demonstrating excellent performance even when the number of labeled samples is minimal. Particularly noteworthy is its performance on DR, where, even at a label rate of only 1%, it outperforms BERT at a 30% label rate. The model exhibits remarkable capability in enhancing DR. On the PeMS dataset, when the number of labeled data is the smallest, only 50, the proposed FPMT model achieves a detection rate 4.4% higher than MixText, demonstrating the best performance. On the I-880 dataset, the detection rate is 5.6% higher than MixText. The improvement on the other two datasets is not as significant, but the model still exhibits the best performance.

**Table 1.** Performance (Classification Rate (CR) (%), Detection Rate (DR) (%), and F1-score (%)) comparison with baselines. Models are trained with 50, 100, and 1500 labeled data per class.

| Dataset | Model | 50 | 500 | 1500 |
|---|---|---|---|---|
| PeMS | BERT | 71.3/63.1/70.6 | 83.8/76.1/81.3 | 89.3/85.8/89.6 |
|  | VAT | 89.3/80.1/89.2 | 91.9/84.1/90.4 | 92.3/84.8/90.9 |
|  | UDA | 88.9/79.1/87.4 | 91.9/83.8/90.1 | 92.4/85.7/90.7 |
|  | DSP | 90.2/80.8/89.9 | 92.1/84.4/90.9 | 92.9/86.8/91.3 |
|  | MT | 91.6/81.9/90.3 | 92.7/85.1/91.7 | 93.1/87.4/92.3 |
|  | FPMT | **93.7/86.3/91.7** | **94.3/87.7/92.8** | **95.5/90.2/94.7** |
| I-880 | BERT | 70.7/64.1/69.6 | 82.8/77.1/81.8 | 88.2/86.8/87.8 |
|  | VAT | 88.7/82.1/87.7 | 90.4/86.5/87.9 | 91.8/87.1/90.1 |
|  | UDA | 88.9/83.8/87.4 | 90.9/85.2/89.8 | 91.9/87.5/91.2 |
|  | DSP | 89.5/82.5/89.2 | 91.4/86.3/91.2 | 92.6/88.4/91.4 |
|  | MT | 90.9/82.7/89.9 | 91.9/87.3/91.6 | 92.9/89.3/92.1 |
|  | FPMT | **93.2/88.3/92.7** | **93.7/89.4/92.9** | **94.9/92.1/94.9** |
| Whitemud Drive | BERT | 84.8/79.4/82.9 | 88.1/83.6/86.8 | 90.7/85.2/89.9 |
|  | VAT | 92.7/87.1/90.4 | 94.5/88.9/92.7 | 95.4/89.2/93.1 |
|  | UDA | 92.4/86.7/90.1 | 93.9/87.6/91.9 | 95.8/89.6/93.7 |
|  | DSP | 93.8/88.1/92.3 | 94.7/89.6/93.4 | 95.9/90.1/94.2 |
|  | MT | 96.7/90.8/94.9 | 98.1/92.8/97.9 | 98.7/92.3/98.2 |
|  | FPMT | **97.3/92.7/95.6** | **98.2/93.1/97.3** | **98.7/93.2/98.4** |
| NGSIM | BERT | 80.8/76.4/82.9 | 85.4/83.6/86.8 | 85.9/83.7/87.1 |
|  | VAT | 89.6/83.4/87.3 | 90.8/85.4/89.9 | 91.3/86.1/90.7 |
|  | UDA | 88.4/82.8/86.5 | 90.5/83.9/88.4 | 91.7/85.3/89.8 |
|  | DSP | 90.5/84.3/88.5 | 91.1/86.2/89.9 | 92.3/87.4/90.7 |
|  | MT | 93.1/87.3/91.2 | 94.7/89.4/93.9 | 95.4/89.9/94.3 |
|  | FPMT | **94.8/90.4/92.2** | **95.9/91.3/93.5** | **96.8/92.3/94.7** |

### 4.5    Ablation Experiments

The experimental results comparing the model PMT of FPMT during the semi-supervised fine-tuning stage with MixText on dataset PeMS and dataset Whitemud Drive are presented in Table 2. It can be observed that, with the improved mixing strategy, the model's performance across various aspects has been enhanced.

Table 2. Performance comparison with MixText and PMT.

| Dataset | Model | 50 | 500 | 1500 |
|---|---|---|---|---|
| PeMS | MT | 91.6/81.9/90.3 | 92.7/85.1/91.7 | 93.1/87.4/92.3 |
|  | PMT | 92.1/83.2/90.9 | 93.9/86.2/92.3 | 94.3/88.5/93.4 |
|  | FPMT | **93.7/86.3/91.7** | **94.3/87.7/92.8** | **95.5/90.2/94.7** |
| Dataset | Model | 50 | 500 | 1500 |
| Whitemud Drive | MT | 96.7/90.8/94.9 | 98.1/92.8/97.9 | 98.7/92.3/98.2 |
|  | PMT | **97.6**/91.3/93.5 | **98.3**/92.9/**97.4** | **98.9**/92.7/98.3 |
|  | FPMT | 97.3/**92.7/95.6** | 98.2/**93.1**/97.3 | 98.7/**93.2/98.4** |

## 5    Conclusion

In the research field of traffic incident detection, methods based on traffic data have made significant progress. However, popular deep-learning approaches heavily rely on data collection and labeling. To alleviate the re-source-intensive nature of data labeling, this paper proposes a semi-supervised learning traffic incident detection model, FPMT, reducing the model's dependence on labeled data. The training pipeline involves pretraining in an unsupervised manner, followed by supervised fine-tuning, and ultimately semi-supervised training. The model incorporates GANs for balancing and augmenting the dataset and utilizes a data augmentation technique, probabilistic pseudo-mixing, at hidden layers to enhance the performance of the semi-supervised model. Comparative experiments with recent models on four real datasets demonstrate the effectiveness of the proposed model. The results show that the model achieves high performance even in scenarios with limited labeled data. As a future research direction, we plan to explore deep semi-supervised learning for traffic incident detection in open environments, simultaneously handling data from different domains or modalities, and leveraging unlabeled data for learning in situations with limited labeled data.

**Acknowledgments** This work is supported by China NSFC Program under Grant NO. 61603257.